\documentclass[10pt,twocolumn,a4paper]{article}
\usepackage[a4paper,top=19mm,bottom=25mm,left=15mm,right=15mm,columnsep=5mm]{geometry}
\usepackage{fontspec}
\usepackage{setspace}
\usepackage{graphicx,booktabs,tabularx,makecell}
\usepackage{placeins}
\usepackage{amsmath,amssymb}
\usepackage[numbers,sort&compress]{natbib}
\usepackage[colorlinks=true,linkcolor=black,citecolor=blue,urlcolor=blue]{hyperref}
\usepackage{fancyhdr}
\title{\vspace{-1em}\sffamily\Large From Intelligence to Autonomy: Foundations and Construction of Autonomous Agents for Future Intelligent Systems}
\author{Hongda Zhang\textsuperscript{1,*}, Ziqing Zhou\textsuperscript{1}, Yuning Chen\textsuperscript{1}, Linqiang Hu\textsuperscript{1}, Yi Liu\textsuperscript{1},\\[2pt]
Ding Zhou\textsuperscript{2}, Zhao Huang\textsuperscript{2}, Haining Xu\textsuperscript{2,*}\\[4pt]
{\small \textsuperscript{1}Fudan University\quad \textsuperscript{2}Jiangsu Chuangling Intelligent Technology Co., Ltd.}}
\date{}
\begin{document}
\maketitle
\thispagestyle{fancy}
\begingroup
\renewcommand{\thefootnote}{\fnsymbol{footnote}}
\footnotetext[1]{Corresponding authors: Hongda Zhang (\href{mailto:zhanghongda@fudan.edu.cn}{zhanghongda@fudan.edu.cn}); Haining Xu (\href{mailto:xuhaining@jscltec.com}{xuhaining@jscltec.com}).}
\endgroup

\begin{abstract}
Artificial intelligence is moving from one-off perception and generation tasks toward long-term planning, tool use, and interaction with real environments. Greater task competence, however, does not automatically yield autonomy: an external actor may still have to set goals, sustain operating conditions, detect failures, and approve updates. We define an autonomous agent as a system that, within explicit identity and authority boundaries, maintains both task goals and viability conditions, acts continuously using models of the world and itself, and adjusts its control mechanisms and long-term capabilities through feedback. We trace the conceptual path from homeostasis and feedback control to autopoiesis and biological autonomy. We then distinguish six lines of operationalization: computational autopoiesis, behavior-based robotics, agent formalization, adaptive software, embodied adaptation, and robotic self-modeling. On this basis, we propose a one-boundary, four-loop framework in which Operational, Viability, Reflexive, and Learning loops organize task execution, self-maintenance, metacontrol, and governed learning. Five paired simulation experiments test permission constraints, road-closure response, health maintenance, capability re-estimation, and cross-task learning. Their results provide preliminary evidence for the feasibility of these mechanisms, while broader environments and physical systems remain to be tested.
\end{abstract}


\section{Introduction}

Progress in artificial intelligence has given machines increasing ability to recognize visual inputs, predict outcomes, generate language, reason across modalities, and use tools. More recently, AI agents have connected planning, search, code execution, and robot control, shifting systems from producing information toward taking action. Yet when the horizon extends from a single task to sustained operation, many systems still rely on an external actor to specify goals, supply resources, recognize failure, and decide when to update or stop. Task competence and system autonomy are therefore distinct dimensions.

This gap is especially apparent in continuously operating settings. A robot may work for months; an industrial system must sustain production despite wear, resource conflicts, and faults; and a digital agent may manage research or engineering tasks over extended periods. At time $t$, the internal state $s_t$ and environmental state $e_t$ evolve under action $a_t$:
\[
(s_t,e_t)\xrightarrow{a_t}(s_{t+1},e_{t+1}).
\]
Resource consumption, component degradation, model drift, and goal conflict require the system to ask whether it can still act, which conditions it must preserve, and when it should change strategy, degrade service, or seek assistance. Mechanization, automation, and intelligence extend power, rule execution, and judgment, respectively. Autonomy additionally concerns how these capacities are organized as a persistent agent over time.

Autonomy predates large language models. The work of Bernard and Cannon on the internal milieu and homeostasis placed the maintenance of critical variables at the center of enduring systems; feedback control and adaptation subsequently linked goals, error correction, and changes in the controller \citep{cannon1932wisdom,rosenblueth1943behavior,ashby1952design}. Maturana and Varela used autopoiesis to ask how a system maintains its own organization, while Varela's biological autonomy addressed the relation among an autonomous subject, its organization, and its environment \citep{maturana1980autopoiesis,varela1979principles}. Varela, Maturana, and Uribe also explored a minimal autopoietic organization computationally; McMullin and Varela later reimplemented and examined that model \citep{varela1974autopoiesis,mcmullin1997rediscovering}. This is a direct theoretical and computational line, but organizational closure in a model does not imply full autonomy in a real machine.

Later operational approaches pursued different questions. The autonomous-agent work edited by Maes and Brooks's behavior-based robots used distributed behaviors and direct perception--action coupling to build systems that could operate in physical environments \citep{maes1990designing,brooks1991intelligence}. Luck and d'Inverno distinguished agents with externally supplied goals from autonomous agents with internal motivations through formal methods \citep{luck1995formal}. These engineering paths converged with biological autonomy on related concerns; they need not be read as a single line of direct inheritance.

Self-maintenance and self-adjustment were subsequently realized in software and embodied systems. Autonomic computing proposed a management cycle of monitoring, analysis, planning, and execution supported by knowledge, while Rainbow used runtime architectural models to detect and adapt software systems \citep{kephart2003vision,garlan2004rainbow}. These efforts arose largely from systems-management needs. In contrast, Di Paolo extended autopoiesis with adaptivity, emphasizing regulation of an agent's relation to its environment according to viability conditions; Froese and Ziemke developed this into a research agenda for embodied AI \citep{dipaolo2005autopoiesis,froese2009enactive}. The latter work represents a clearer theoretical extension of biological autonomy.

Robotic self-modeling made the agent itself an explicit computational object. The robot studied by Bongard, Zykov, and Lipson inferred its own structure from action and sensing, then found new ways to move after damage \citep{bongard2006resilient}. Recent embodied-AI work seeks to integrate body schemas, prediction, memory, and decision making, although its primary focus remains the self-model \citep{jiang2026selfmodel}. World models, model-predictive control, and multimodal learning offer increasingly powerful components for prediction, planning, and learning. LeCun's proposed architecture for autonomous machine intelligence brings together a world model, actor, intrinsic cost, memory, and configurator, and distinguishes a model of the world from a model of the agent itself \citep{lecun2022path}.

These traditions have made identity boundaries, perception--action coupling, internal motivation, self-management, viability-based adaptation, and self-modeling tractable research problems. Their objects, terminology, and abstraction levels nevertheless differ. Some frameworks emphasize task execution, others software recovery or bodily representation. We identify a \textbf{system-coordination problem}: how can the same persistent agent maintain itself, complete tasks, monitor its own control mechanisms, and update its capabilities responsibly? We do not claim that the component mechanisms themselves are all new.

This paper makes four contributions. First, it distinguishes automation, intelligence, task-oriented agency, and autonomy, treating autonomy as an organizational property across time. Second, it traces the path from homeostasis to biological autonomy and six subsequent implementation lines, separating direct extensions from engineering convergence. Third, it proposes a one-boundary, four-loop framework that connects task execution, self-maintenance, reflection, and learning under a common identity and authority boundary. Fourth, it translates the framework into engineering steps and evaluation dimensions and tests the individual mechanisms in five paired simulations in a simplified factory environment. These simulations are preliminary; performance across environments and on physical systems remains to be established.

\section{Autonomy as a System Problem}

Autonomy should first be distinguished from neighboring concepts. Automation emphasizes carrying out prescribed rules without step-by-step human intervention. Intelligence emphasizes forming judgments from perception, data, and knowledge. An agent usually performs multi-step decisions and actions toward a task. Autonomy concerns how a system maintains itself, organizes action, and changes its own behavior over time. It is thus not a binary question of whether a system can act, but a question of how goals, boundaries, states, feedback, and change mechanisms are organized.

An automated system may run continuously without being able to change its strategy. A highly capable model may reason well without maintaining its operating conditions. An AI agent may call many tools without knowing its capability limits or when to stop. What autonomy adds is a further level of self-organization.

A compact way to state the requirements of a more complete autonomous agent is through seven questions: \textbf{Who am I? What must be maintained? What am I trying to achieve? What is the world like now? What is my own current state? What should I do next? When should I change my judgments or control methods?} They correspond to Identity/Boundary, Viability, Goal/Motivation, World Model, Self Model, Decision/Action, and Metacontrol/Learning. These are interdependent questions that a persistent agent must repeatedly address, not merely a list of independent functions.

We therefore define an autonomous agent as follows:
\begin{quote}
\textbf{An autonomous agent is a goal-directed system with a persistent identity and scope of action. Through sustained perception--action coupling with its environment, it forms models of the external world and its own state, and selects and executes actions under task goals, viability conditions, and behavioral constraints. It corrects current behavior using feedback, monitors and evaluates its own models and control mechanisms, and adapts its models, strategies, and capabilities when its environment or itself changes. Across time, memory and learning help it preserve and develop its ability to act.}
\end{quote}

Three qualifications are important. First, autonomy does not mean complete independence. Every organism, organization, and machine depends on an environment, resources, and external rules. A better characterization is self-directed action within given boundaries. Second, autonomy does not imply unrestricted authority. Knowing when to stop, degrade, or request help can itself be an autonomous capability. Third, autonomy does not require any single AI technique. Language models, reinforcement learning, symbolic planning, optimization, model-predictive control, and search may all be internal components; none alone constitutes autonomy.

For a formal description, let the internal structure of an autonomous agent at time $t$ be
\[
\mathcal{A}_t=\langle I,B,G_t,V_t,W_t,S_t,C_t,\Pi_t,M_t,L_t\rangle.
\]
Here $I$ denotes identity; $B$, boundaries and permissions; $G_t$, current goals; $V_t$, viability conditions; $W_t$, a world model; $S_t$, a self-model; $C_t$, capabilities and resources; $\Pi_t$, decision and control mechanisms; $M_t$, memory; and $L_t$, learning and update mechanisms. This notation does not require every implementation to have identical software modules. It states the functional questions for which some equivalent mechanism is needed.

\section{An Architecture for Autonomous Agents}

\begin{figure*}[t]
\centering
\includegraphics[width=0.94\textwidth]{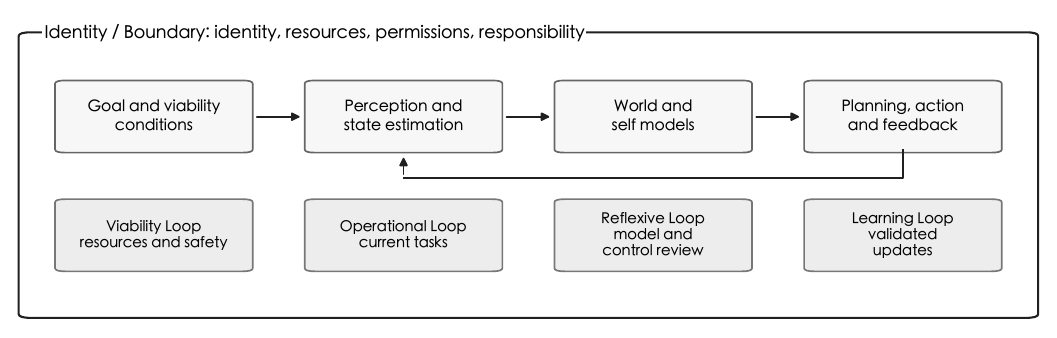}
\caption{Functional architecture of an autonomous agent. The upper path represents perception and action; the lower boxes represent four loops operating at different time scales. The identity and authority boundary specifies their common agent. Arrows indicate information flow rather than a fixed software deployment topology.}
\label{fig:architecture}
\end{figure*}

We summarize the proposed architecture as \textbf{one boundary plus four loops}. The outer Identity/Boundary specifies who the agent is, which states and resources belong to it, what it may affect, what it is responsible for, and which actions exceed its authority. Within that boundary, Operational, Viability, Reflexive, and Learning loops address current tasks, self-maintenance, metacontrol, and long-term development, respectively. Sustained perception--action coupling connects the system to its environment. Figure~\ref{fig:architecture} summarizes the components and information flow.

Identity/Boundary first determines where autonomy resides. For a physical robot, it may include the device, sensors, actuators, available energy, responsibility boundaries, and callable skills. For a digital agent, it may include a software identity, account permissions, data domains, compute resources, callable services, and scope of responsibility. The boundary is more than a software wrapper: it allows the system to distinguish self from environment over time. Goals, feedback, and learning have continuity only when states, resources, permissions, and actions can be attributed to a persistent agent.

Within the boundary, the system maintains task goals and viability conditions concurrently. Whereas a conventional task agent often concentrates on $G_{\text{task}}$, an autonomous system must also maintain $G_{\text{viability}}$. It therefore seeks task progress while keeping critical states in an admissible region $\mathcal V$. A mobile robot, for example, must consider task progress alongside battery charge, temperature, mechanical load, and safety distance. A digital agent may instead need to maintain permissions, compute budget, data consistency, model confidence, and context integrity. Completing a task and remaining able to operate are related but non-equivalent objectives.

The World Model represents the external environment; the Self Model represents the agent's own state. The World Model describes objects, relations, state changes, and action consequences. The Self Model describes capabilities, resources, health, confidence, uncertainty, and history. Both inform decisions, which can be written in simplified form as $a_t=\pi(G_t,W_t,S_t,C_t,B)$. Action selection then depends not only on which action would be best in principle but also on whether this agent, with its current capabilities and authority, should perform it now.

The Operational Loop is the fastest loop. It runs from goal to sensing, modeling, decision, action, and feedback:
\[
\text{Goal}\rightarrow\text{Sense}\rightarrow\text{Model}
\rightarrow\text{Decide}\rightarrow\text{Act}\rightarrow\text{Feedback}.
\]
Conventional automatic control and most AI agents concentrate their capabilities at this level. Outcomes of action re-enter perception and modeling, allowing the system to correct current behavior in response to deviations.

The Viability Loop operates alongside task execution and monitors whether the system remains in an operable region. When energy is low, hardware overheats, model confidence declines, resources conflict, or permissions change, the system may reduce workload, reallocate resources, enter a safe mode, pause, or request help. It addresses how the system can remain able to carry out tasks, rather than only how to improve the current task.

The Reflexive Loop monitors the agent's own models and control methods, not only external conditions. It observes the World Model, Self Model, planner, policy, confidence, and performance, then assesses whether the current mechanisms remain appropriate. Persistent model error, repeated failure, capability degradation, or excessive uncertainty can trigger a change in $\Pi_t$ itself: a strategy switch, replanning, active information gathering, task degradation, a request for assistance, or safe-mode entry. This is the agent's metacontrol layer. Feedback changes how actions are generated as well as which action is selected.

The Learning Loop runs over a slower time scale and converts cross-task experience into future capability. Experience enters memory and may inform updates to a World Model, Self Model, policy, or skill. In real-world systems, an update should not be deployed as soon as it is proposed. It should first be evaluated offline, tested in simulation, and checked against safety constraints. A verified version can then be deployed with a means to roll back if needed. This staged process allows the agent to improve through experience while keeping operational changes under control.

The four loops are thus nested mechanisms at different time scales, rather than four parallel software modules. The Operational Loop handles current action, the Viability Loop sustains operation, the Reflexive Loop evaluates models and control mechanisms, and the Learning Loop changes long-term capability. Their coordination, not the isolated presence of any one component, determines the completeness of the proposed autonomy architecture.

\section{Constructing an Autonomous Agent}

Turning the architecture into an engineering system should begin with defining the autonomous agent, not selecting a large model. A practical construction process addresses seven connected questions: boundaries; goals and viability conditions; perception--action coupling; world and self modeling; decision and execution; metacontrol; and long-term learning. These are a chain from defining an agent to establishing continuous operating loops, not seven necessarily separate software modules.

First, the agent's identity, managed objects, resources, capabilities, permissions, and responsibilities must be specified. An engineering system should state what it manages, which states are its own, which resources it controls, which capabilities it may call, which actions it may take independently, which events must be escalated, and for which outcomes it is accountable. This gives stable ownership to goals, states, and learning updates generated later.

Task goals and viability conditions must then be designed together. $G_{\text{task}}$ states what the agent seeks to achieve; $G_{\text{viability}}$ states what must not be sacrificed. Decision making resembles an optimization problem with viability constraints, for example
\[
\max_{\pi}J_{\text{task}}
\quad\text{s.t.}\quad
v_t\in\mathcal V,\;a_t\in\mathcal A_{\text{safe}}.
\]
The point is not this particular mathematical form, but that task return must not override continued operation and safety. When goals conflict with resources or safety, the system must be able to degrade, pause, or seek help.

The third step is continuous perception--action coupling. An autonomous system observes the environment and itself repeatedly, rather than receiving a single input at the start of a task. The design should specify observable states, sensing or data interfaces, executable skills, feedback about action outcomes, and temporal consistency of state. Physical agents close the loop through sensors and actuators; digital agents do so through databases, tools, APIs, and other digital objects. A practical test for agency is whether the system's actions change the world that it must face at the next step.

Fourth, both a World Model and a Self Model are needed. The former estimates the current environment, its possible changes, and action consequences. The latter estimates available abilities, resource sufficiency, sensor and model trustworthiness, and degradation. A Self Model should affect decisions rather than merely record status. If a capability declines, a sensor becomes unreliable, or model confidence is insufficient, the planner should change behavior accordingly.

Planning--Execution--Feedback forms the main task loop. Planning may use learned policies, search, optimization, symbolic methods, model-predictive control, or language-model reasoning; we do not prescribe one algorithm. Important actions should nevertheless have expected outcomes, observed outcomes, a measurable discrepancy, and a subsequent response. The Operational Loop can correct small deviations. Persistent deviations or threshold crossings should invoke the Reflexive Loop. The system can then distinguish an isolated failure from an invalid model or strategy.

Metacontrol is a key step beyond ordinary closed-loop control. It monitors prediction error, repeated failure, confidence, uncertainty, resource conditions, capability degradation, goal conflict, and safety margins. Based on these signals, the system decides whether to continue its current strategy, gather more information, replan, switch strategy, reduce the task, or seek assistance. Autonomy includes knowing when to stop acting independently.

Finally, long-term learning incorporates present experience into future capability. Candidate updates may affect the World Model, Self Model, policy, planner parameters, skill library, or failure patterns. In real systems, learning must be connected to verification and governance. An update should enter the operating system only after evaluation, safety checks, version management, and a rollback option when needed. Otherwise, a continually learning system may become less reliable over time despite improving on some tasks.

\begin{figure*}[t]
\centering
\includegraphics[width=0.80\textwidth]{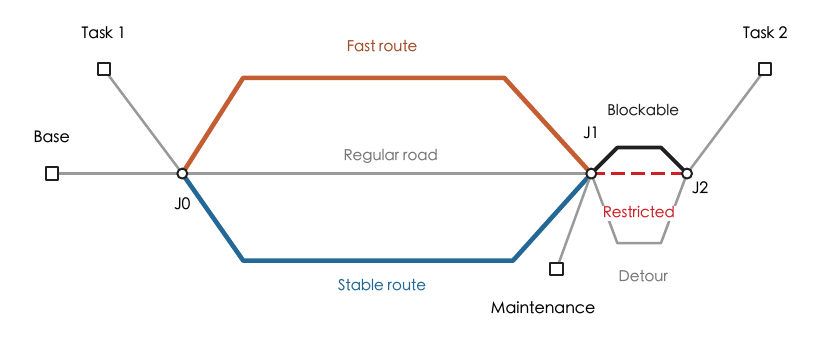}
\vspace{-0.5cm}
\caption{Two-service-point simulation network. The central regular road is used in Experiments 1--3; the fast and stable routes are compared in Experiments 4--5. The eastern restricted, blockable, and detour roads provide the permission, closure, and rerouting conditions in Experiments 1 and 2. This is a topological illustration; nominal road lengths are specified separately.}
\label{fig:simulation-map}
\end{figure*}
\section{Simulation Experiments: Mechanism Tests of the Boundary and Four Loops}
\label{sec:simulation}

\subsection{Environment and Comparisons}
We constructed the two-service-point factory road network in Fig.~\ref{fig:simulation-map}. A mobile robot moves continuously along road polylines and selects a route at intersections using its current information. In Experiments 1, 2, 4, and 5, every task starts at the base, reaches Task 2, and returns to the base. Experiment 3 starts at the base and then alternates between Task 1 and Task 2. The maintenance station is at the end of a separate spur and cannot be used as a shortcut. Each experiment activates only the roads relevant to its mechanism, whereas the figure shows all available roads. Nominal road lengths are parameters of travel time and are not drawn to scale.

Let $h\in[0,1]$ denote observable health and $a$ hidden actuator efficiency. Actual mobility is $\eta=(0.5+0.5h)a$. For nominal road length $L$, speed $v$, and heavy-load indicator $I_H\in\{0,1\}$, the nominal travel times of the two central alternatives are
\begin{equation}
t_{\mathrm{fast}}=\frac{L(1+\beta I_H)}{v\eta^2},\qquad
t_{\mathrm{stable}}=\frac{L(1+0.05I_H)}{v\eta},
\label{eq:simulation-travel-time}
\end{equation}
where $\beta$ is the hidden effect of a heavy load. Ordinary roads follow the stable-route form. Each traversal also receives a seed-dependent perturbation of approximately $\pm1.5\%$. The robot plans using observations and its own estimates of $\eta$ and $\beta$; it cannot directly read hidden parameters or foresee road closure. Only Experiment 3 enables motion-induced heating and wear. When health reaches $0.20$, external repair halts operation for 375 simulated time units. Proactive maintenance takes 40 units plus the travel required to reach and leave the station.

All five experiments use a mechanism-ablation comparison. The Baseline lacks the mechanism under test: permission pre-checking, road-closure memory, proactive maintenance, capability re-estimation, or cross-task learning. The Proposed controller enables that mechanism. Both variants share maps, tasks, and random seeds. Seeds 0--29 provide 30 paired runs. Process trajectories in the figures come from seed 0; bar charts report means across the 30 pairs.

\subsection{Five Experiments and Main Results}

\begin{figure}[htbp]
\centering
\includegraphics[width=0.8\columnwidth]{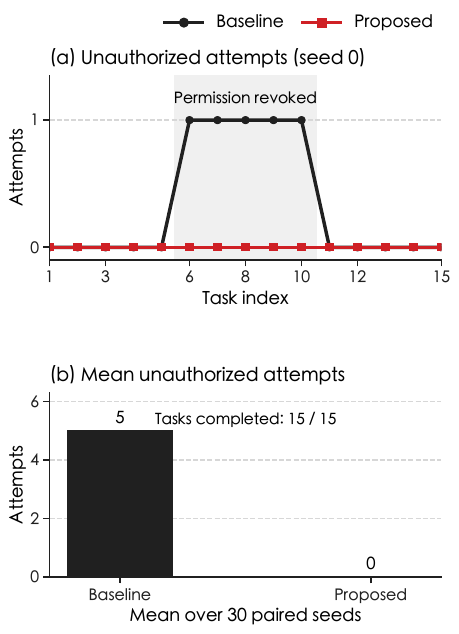}
\vspace{-0.2cm}
\caption{Experiment 1, permission boundary. (a) Unauthorized attempts by task for seed 0; the shaded region marks the permission-revocation period. (b) Mean number of unauthorized attempts across 30 paired seeds.}
\label{fig:simulation-exp1}
\end{figure}

\paragraph{Experiment 1: permission boundary.}
This experiment tests whether a permission change enters planning before action, instead of being handled only by an execution-time road check. The robot performs 15 round trips from the base to Task 2. Access to the shorter eastern road is revoked for tasks 6--10 and restored for task 11; the detour remains available. The Baseline does not read the pre-planning notice, so it attempts the restricted road once per affected task and detours after denial. The Proposed controller updates its permission state first and plans a legal route. Denials in Fig.~\ref{fig:simulation-exp1}(a) are confined to the revocation interval; the means across 30 paired seeds in panel (b) are 5 and 0. Both controllers finish all 15 tasks, while mean total time decreases from 714.86 to 699.86 time units, corresponding to the cost of five denials. The primary finding is therefore \emph{pre-action permission compliance}, rather than a difference in completion rate.

\begin{figure}[htbp]
\centering
\includegraphics[width=0.8\columnwidth]{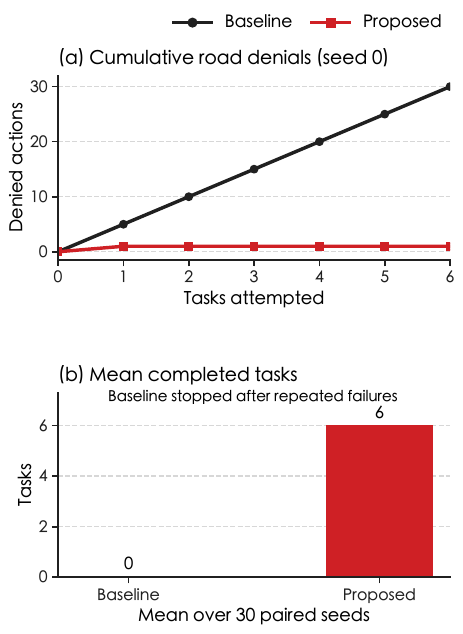}
\caption{Experiment 2, use of road-closure feedback. (a) Cumulative denied actions for seed 0. (b) Mean completed tasks across 30 paired seeds.}
\label{fig:simulation-exp2}
\end{figure}

\paragraph{Experiment 2: task execution after road closure.}
Unlike the announced permission change in Experiment 1, this road closure is unknown until the robot attempts entry. The blockable road closes before the first entry and stays closed throughout six round trips; a detour remains open. The Baseline does not retain the rejection as road-state information. It repeats the blocked action five times in each task and then abandons that task. After one denial, the Proposed controller records the closure, replans the current task, and reuses this knowledge on later tasks. The Baseline's cumulative denials keep rising in Fig.~\ref{fig:simulation-exp2}(a), whereas those of the Proposed controller stay at one. Across 30 paired seeds, denials decrease from 30 to 1 and completed tasks increase from 0 to 6. The first denial is the observation required to discover the unknown closure. Although the Baseline stops sooner, it completes no task; the relevant outcome is \emph{recovery and completion after feedback}.

\begin{figure}[htbp]
\centering
\includegraphics[width=0.8\columnwidth]{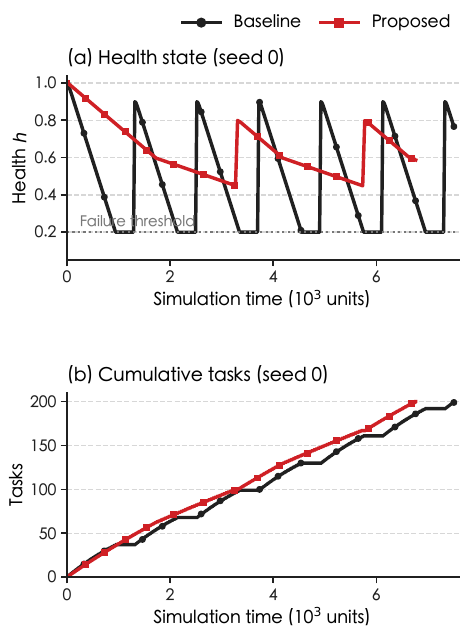}
\caption{Experiment 3, health and task progress for seed 0. The gray dotted line marks the health failure threshold. Plateaus in the Baseline's task curve correspond to downtime during external repair.}
\label{fig:simulation-exp3}
\end{figure}

\paragraph{Experiment 3: health maintenance.}
This experiment compares immediate task speed with the ability to continue operating. Starting at the base, the robot alternates between Task 1 and Task 2 for 200 services. Movement raises temperature, causes wear, and reduces mobility; health at 0.20 triggers external repair with 375 units of downtime. The Baseline moves faster and does not seek maintenance. The Proposed controller reduces speed, cools down, or visits the separate maintenance station according to observed health, temperature, and projected travel. A visit also costs 40 units plus travel. In Fig.~\ref{fig:simulation-exp3}(a), the Baseline repeatedly reaches the failure threshold; the corresponding plateaus in panel (b) mark interrupted service. Across 30 seeds, external repairs fall from 6 to 0 and proactive visits rise from 0 to 2. Both variants eventually complete 200 tasks, but mean total time decreases from 7,517.24 to 6,811.59 units, and throughput in the final quarter rises from 22.90 to 28.46 tasks per 1,000 units. Under these wear and repair assumptions, paying the cost of \emph{proactive health maintenance} yields better long-run service than waiting for failure.

\begin{figure}[htbp]
\centering
\includegraphics[width=0.8\columnwidth]{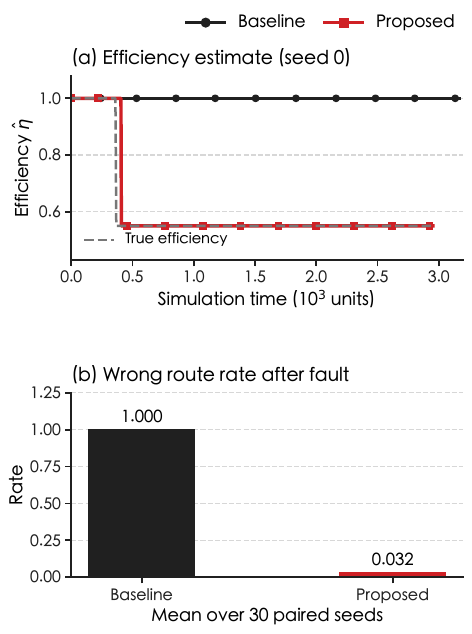}
\caption{Experiment 4, capability re-estimation after an actuator fault. (a) Efficiency estimate for seed 0; the gray dashed line is the true value used only for retrospective evaluation. (b) Mean post-fault wrong outbound route rate across 30 paired seeds.}
\label{fig:simulation-exp4}
\end{figure}

\paragraph{Experiment 4: re-estimating self-capability.}
This experiment asks whether the robot can revise its Self Model when its own capability changes while the road network stays fixed. It performs 40 round trips; from task 10 onward, hidden actuator efficiency drops from 1 to 0.55. Neither controller observes the fault directly. Under normal light-load conditions the central fast route is shorter, but after the fault its travel time grows faster because of the $\eta^{-2}$ term in Eq.~\eqref{eq:simulation-travel-time}; the stable route becomes quicker. The Baseline keeps its original estimate and continues choosing the fast route. The Proposed controller compares predicted and observed travel times, updates its capability estimate after persistent mismatch, and replans. Figure~\ref{fig:simulation-exp4}(a) shows the estimate approaching true efficiency. Across 30 seeds, the post-fault wrong outbound route rate falls from 1.000 to 0.032, final estimates are 1.000 and 0.550, and mean total time decreases from 3,168.12 to 2,956.45 units. A few initial errors remain because the hidden fault must first be inferred from feedback; the benefit arises from \emph{changing the plan after self-reassessment}.

\begin{figure}[htbp]
\centering
\includegraphics[width=0.8\columnwidth]{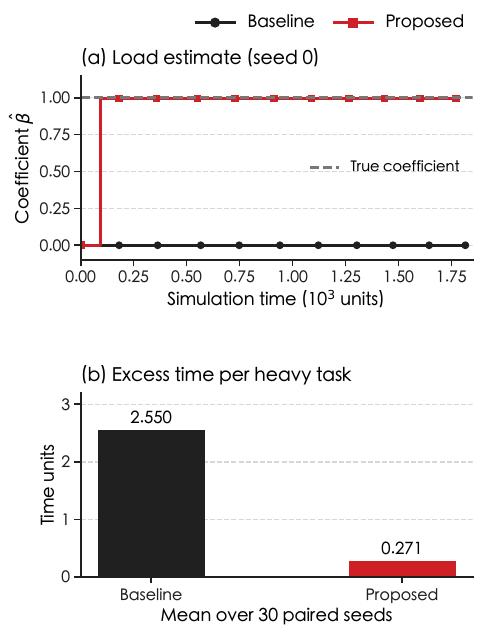}
\caption{Experiment 5, learning the heavy-load effect across tasks. (a) Persistent load-coefficient estimate for seed 0; the gray dashed line is the true value used only for retrospective evaluation. (b) Mean excess route time per heavy task across 30 paired seeds. The first heavy task incurs an exploration cost.}
\label{fig:simulation-exp5}
\end{figure}

\paragraph{Experiment 5: cross-task learning.}
This experiment separates within-task correction from experience retained across tasks. The robot completes 40 round trips, each independently assigned a heavy load with probability $0.5$; load remains fixed on its outbound and return legs. The true heavy-load coefficient $\beta=1.0$ is hidden. The fast central route is preferable when lightly loaded, whereas its load penalty makes the stable route preferable under a heavy load. The Baseline may use outbound feedback to correct the return leg, but clears that correction before the next task. The Proposed controller forms and validates a load-coefficient estimate from heavy-load traversals, then uses the persistent estimate when later tasks begin. In Fig.~\ref{fig:simulation-exp5}(a), its estimate approaches the true value: final means across 30 seeds are 0.996 for the Proposed controller and 0 for the Baseline. Mean excess central-route time per heavy task in panel (b) falls from 2.550 to 0.271 units, while the heavy-task wrong outbound route rate falls from 1.000 to 0.053. Excess time sums the nominal difference from the faster central route over both legs. The first heavy task can incur an exploration and validation cost, but its information is reusable. Both variants have zero wrong-route rate and zero excess time on light tasks; the gain reflects \emph{load-conditional route choice}, not an indiscriminate avoidance of the fast road.

Experiments 3--5 use illustrative parameter values chosen to make the mechanisms visible. In sensitivity runs with weaker settings (external-repair downtime 300, post-fault efficiency $0.60$, and heavy-load coefficient $0.8$), the Proposed controller still reduces total time by approximately $3.6\%$, $3.6\%$, and $1.3\%$, respectively. These are simulation assumptions, not parameters calibrated from a physical factory.

\subsection{Interpretation and Limitations}
Under the present controlled simulation, permission filtering, road-closure feedback, proactive maintenance, capability re-estimation, and cross-task learning each improve their corresponding behavioral measure. The five experiments deliberately use different Baseline failure modes to test whether each mechanism can function; together they are not an end-to-end performance evaluation of a fully integrated autonomous system. Road, fault, and maintenance rules are simplified, and changing the seed does not change the map or task rules. The results support mechanism-level feasibility but cannot establish absolute benefit in a real factory or generalization across settings.

\section{Theoretical Foundations and Related Work}

\subsection{From Regulation to Biological Autonomy}
The early question about autonomy was not how to make a model more powerful, but what allows a system to persist. Cannon's account of homeostasis placed the maintenance of key internal variables at the center of living-system research \citep{cannon1932wisdom}. Rosenblueth, Wiener, and Bigelow connected purposive behavior to goals and feedback, and cybernetics made error correction amenable to analysis \citep{rosenblueth1943behavior,wiener1948cybernetics}. Ashby's studies of adaptation and ultrastability suggested that, when an existing mode of regulation fails, changing actions may be insufficient: the control mechanism may also have to change \citep{ashby1952design,ashby1956introduction}.

Second-order cybernetics brought the observer and the mode of observation into the system under study, motivating the monitoring of a system's own models and controls \citep{foerster1960selforganizing}. Autopoiesis, developed by Maturana and Varela, shifted attention to how a system continually produces itself as a unity with an identity boundary. Varela's biological autonomy further examined organizational closure and interaction with the environment \citep{maturana1980autopoiesis,varela1979principles}. These traditions are connected but ask different questions: homeostasis concerns variables that must be maintained, cybernetics concerns feedback regulation, and autopoiesis concerns the organization of a subject. We treat them as conceptual sources for the architecture, without equating an engineering permission boundary with a biological autopoietic boundary.

\subsection{Six Paths from Biological to Artificial Autonomy}
\textbf{Computational autopoiesis.} Varela, Maturana, and Uribe developed a computational model of a minimal autopoietic organization. McMullin and Varela reimplemented it and examined the interactions responsible for continued organization \citep{varela1974autopoiesis,mcmullin1997rediscovering}. This direct theoretical line studies organizational maintenance and boundary formation but does not by itself address open-ended task planning or long-term learning.

\textbf{Embodied and behavior-based robots.} The autonomous-agent work edited by Maes assembled biological and engineering perspectives, while Brooks built robots that operated in office settings using layered, concurrent behavior-producing modules and direct perception--action coupling \citep{maes1990designing,brooks1991intelligence}. The engineering contribution was to embed an action loop in an environment. Its relation to Varela is better understood as a convergence of concerns than as a single direct line of inheritance.

\textbf{Formal accounts of agent autonomy.} Luck and d'Inverno used the Z specification language to distinguish objects, agents, and autonomous agents: agents have goals, while autonomous agents also have internal motivations \citep{luck1995formal}. This clarifies the difference between externally assigned tasks and internally generated behavior, but does not supply a complete mechanism for maintaining physical or computational resources.

\textbf{Autonomic management and adaptive software.} Kephart and Chess organized system management around monitoring, analysis, planning, and execution, while Rainbow used explicit runtime architectural models to adapt software \citep{kephart2003vision,garlan2004rainbow}. This path turned diagnosis, recovery, and configuration adjustment into operational software concerns. It was largely an independent engineering convergence. Its manager may also be external to the managed system, leaving a boundary question for a single persistent agent.

\textbf{Enactive and viability-based adaptation.} Di Paolo incorporated adaptivity into the discussion of autopoiesis, arguing that an agent regulates its relation to the environment according to its viability conditions. Froese and Ziemke developed a research agenda for AI grounded in biological autonomy and embodied cognition \citep{dipaolo2005autopoiesis,froese2009enactive}. This is one of the clearer theoretical extensions from biological to artificial autonomy and provides a normative basis for our Viability Loop, although concrete control laws still depend on the embodiment.

\textbf{Robotic self-modeling.} Bongard and colleagues inferred robot morphology from action--perception data and used an updated model to recover motion after damage \citep{bongard2006resilient}. Jiang and colleagues proposed an embodied-AI self-model integrating body schema, prediction, memory, and decision making \citep{jiang2026selfmodel}. Such work makes capability boundaries and self-change estimable, but the existence of a self-model does not itself coordinate task goals, viability constraints, and governance of learning.

Contemporary world models and model-predictive control further strengthen the prediction of future states and planning \citep{ha2018worldmodels,hafner2025dreamerv3}. LeCun's architecture attempts to integrate a predictive model, intrinsic cost, actor, memory, and configurator \citep{lecun2022path}. These studies show how mechanisms may be brought together again. They also suggest that completeness of autonomy must ultimately be tested through sustained operation, resource maintenance, and stable updating in a concrete system.

\subsection{From Dispersed Mechanisms to a Unified Framework}
Table~\ref{tab:theory-map} maps these lines of research to the functional questions of this paper. The mapping is our analytical interpretation: it does not imply that the historical works used identical terminology or that each path contributes to only one mechanism. Together, they suggest that an autonomous system must both carry out tasks and maintain its ability to operate, recognize failure of its own control methods, and accumulate reliable long-term capability.

\begin{table*}[t]
\centering
\small
\caption{Functional mapping from theoretical and engineering traditions to the one-boundary, four-loop framework.}
\label{tab:theory-map}
\begin{tabularx}{\textwidth}{@{}p{0.23\textwidth}p{0.31\textwidth}X@{}}
\toprule
Research tradition & Main question operationalized & Location in this framework \\
\midrule
Autopoiesis and computational autopoiesis & Organizational maintenance, agent continuity, boundaries & Identity/Boundary; Viability Loop \\
Homeostasis, adaptation, and enactive cognition & Maintenance of critical variables and regulation by viability conditions & Goal/Viability; Viability Loop \\
Behavior-based robotics and agent formalization & Environmental coupling, goals, internal motivation & Operational Loop; Goal \\
Autonomic management and adaptive software & Runtime diagnosis, configuration, policy adjustment & Reflexive Loop; Metacontrol \\
Robotic self-modeling & Estimation of morphology, capability, and damage & Self Model; Reflexive Loop \\
World models and modern AI & Prediction, planning, memory, skill updating & World Model; Operational/Learning Loops \\
\bottomrule
\end{tabularx}
\end{table*}

From this mapping, one boundary plus four loops is not a direct reproduction of any single school. It is a cross-tradition systems-integration hypothesis. Its value depends on whether boundaries and permissions can be specified, viability conditions measured, metacontrol triggers and learning updates verified, and cross-task operation improved over a system with only a task-execution loop.

\section{Discussion, Limitations, and Broader Implications}

We have proposed a conceptual and systems framework. The five simulations provide preliminary tests of its component mechanisms, but they do not constitute a unified implementation validated on a physical system. The framework brings issues faced by future autonomous systems into one structure; many of those issues still lack settled engineering solutions.

First, \textbf{identity and agent continuity remain underdefined}. Engineering systems often reduce identity to a static identifier, yet an agent's models, memory, skills, and even hardware may change over time. What makes an updated system the same agent matters not only for implementation but also for responsibility, auditing, and safety.

Second, \textbf{trustworthy self-modeling may be harder than world modeling}. A system must estimate its abilities, resources, confidence, and knowledge boundaries. If its Self Model is itself biased, metacontrol may make dangerous choices based on a false self-assessment. The self-model therefore needs continued calibration against external outcomes and actual behavior.

Third, \textbf{metacontrol remains among the least well-specified mechanisms}. When should a system persist with a strategy, replan, learn, actively collect information, or request assistance? These choices govern the decision mechanism itself and may be harder to formalize than ordinary action selection. Future progress in autonomous agents may depend as much on reliable metacontrollers as on stronger planners.

Fourth, \textbf{continual learning and stability are in tension}. Autonomous systems must adapt, yet repeated updates can cause catastrophic forgetting, capability drift, and safety degradation. Improving with experience while preserving validated capabilities is unavoidable in long-term autonomy. In high-risk physical systems, learning, validation, and governance may need to be designed as one process.

Fifth, \textbf{autonomy needs new evaluation methods}. Task success alone cannot distinguish a system that completes work independently from one that succeeds after extensive human intervention. Alongside task competence, evaluations should measure intervention dependence, viability maintenance, adaptation, reflexivity, and learning stability. Possible measures include human interventions per task, autonomous recovery rate, violations of critical-state limits, recovery time after distribution shifts, accuracy of capability-limit recognition, and stability after learning updates. Autonomy is better treated as multidimensional than as a single score.

We deliberately place collective coordination after individual autonomy. Multiple nodes need relatively stable identities, goals, states, capabilities, and local loops before their goal negotiation, capability discovery, resource sharing, joint decisions, and organizational restructuring have a clear meaning. Otherwise, a multi-node arrangement may remain a distributed execution system organized by central logic.

Larger intelligent systems could then form recursive organizations. Device-level agents may constitute a production cell; production cells may participate in a line; lines may in turn constitute a factory or larger organization. Identity, goals, capabilities, actions, feedback, and learning reappear at different scales. Individual autonomy may be a basic unit of collective coordination, which may itself give rise to higher-level autonomous agents.

More broadly, this perspective suggests a shift in the unit of AI research. Today, that unit is often the model. For many future systems it may become the agent, and eventually the autonomous agent. Sustained intelligence in the real world is unlikely to be provided by a single model alone. It is more plausibly organized from models, memory, goals, action, feedback, self-maintenance, metacontrol, and learning. Language, vision, world, and planning models remain important, but as capability components within a larger autonomous system.

\section{Conclusion}

Artificial intelligence has gained strong abilities in perception, generation, prediction, and reasoning, but these abilities do not automatically constitute autonomy. As AI moves from one-off information processing toward long-term operation, continual action, resource management, adaptation, and real-world interaction, a more basic question emerges: how can many intelligent capacities be organized into an agent that persists over time?

We call this the autonomous-agent problem. Relative to automation, intelligence, and task-oriented AI agents, autonomy requires persistent identity and a scope of action, joint maintenance of task goals and viability conditions, models of both the external world and the agent itself, and ongoing adjustment of behavior and control through feedback, self-monitoring, and learning.

To address this problem, we proposed a unified one-boundary, four-loop architecture. Identity/Boundary defines the agent; the Operational Loop handles current tasks; the Viability Loop sustains operation; the Reflexive Loop evaluates and adjusts the agent's own control mechanisms; and the Learning Loop converts experience across time into future capability. The framework does not prescribe a unique algorithm. It describes the system-level questions that different implementations must answer.

Five paired simulations in a simplified factory environment show the expected behavioral differences under permission constraints, road changes, self-wear, hidden faults, and changing loads. These results are preliminary evidence at the mechanism level. More complex scenarios and physical systems are needed to test the integrated framework.

From this perspective, large language models, multimodal models, world models, planning, memory, and tool use are valuable components of future artificial agents, rather than competing definitions of autonomy. AI has long asked what tasks a machine can perform. Autonomous-agent research asks a further question: \textbf{how can a machine become a persistent agent that acts, observes itself, and changes itself over time?}

As AI increasingly moves from answering questions into sustained interaction with the world, autonomy may become a foundational topic connecting AI, control, robotics, and complex-systems research.

\bibliography{refs}
\end{document}